# BHAAV (भाव) - A Text Corpus for Emotion Analysis from Hindi Stories

**Yaman Kumar**
Adobe Systems, Noida
ykumar@adobe.com

**Debanjan Mahata**[*]
Bloomberg LP
dmahata@bloomberg.net

**Sagar Aggarwal**
NSIT-Delhi
sagara.co@nsit.net.in

**Anmol Chugh**
Adobe Systems, Noida
achugh@adobe.com

**Rajat Maheshwari**
USICT, New Delhi
rajat.usict.101164@ipu.ac.in

**Rajiv Ratn Shah**
IIIT-Delhi
rajivratn@iiitd.ac.in

## Abstract

In this paper, we introduce the first and largest Hindi text corpus, named BHAAV (भाव), which means *emotions* in Hindi, for analyzing emotions that a writer expresses through his characters in a story, as perceived by a narrator/reader. The corpus consists of 20,304 sentences collected from 230 different short stories spanning across 18 genres such as प्रेरणादायक (Inspirational) and रहस्यमयी (Mystery). Each sentence has been annotated into one of the five emotion categories (*anger*, *joy*, *suspense*, *sad*, and *neutral*), by three native Hindi speakers with at least ten years of formal education in Hindi. We also discuss challenges in the annotation of low resource languages such as Hindi, and discuss the scope of the proposed corpus along with its possible uses. We also provide a detailed analysis of the dataset and train strong baseline classifiers reporting their performances.

## 1 Introduction

Emotion analysis from text is the study of identifying, classifying and analyzing emotions (*e.g.*, *joy*, *sadness*), as expressed and reflected in a piece of given text (Yadollahi et al., 2017). Its wide range of applications in areas such as *customer relation management* (Bougie et al., 2003), *dialogue systems* (Ravaja et al., 2006), *intelligent tutoring systems* (Litman and Forbes-Riley, 2004), *analyzing human communications* (Kövecses, 2003), *natural text-to-speech systems* (Francisco and Gervás, 2006), *assistive robots* (Breazeal and Brooks, 2005), *product analysis* (Knautz et al., 2010), and *studying psychology from social media* (De Choudhury et al., 2013), has drawn considerable attention from the scientific community making it one of the important areas of research in computational linguistics.

Majority of the methods and resources developed in emotion analysis domain deals with English language (Yadollahi et al., 2017). Moreover, since there are no text based resources for emotion analysis in Hindi, our understanding of expression of emotions is only limited to English text. To this end, we develop a text corpus[1] for emotion analysis from stories written in Hindi, which is one of the 22 official languages of India and is among the top five most widely spoken languages in the world[2]. The proposed corpus is the largest annotated corpus for studying emotions from Hindi text, and facilitates the development of linguistic resources in low-resource languages.

According to a joint report by KPMG and Google[3] published in 2017, there are 234 million Internet users in India using one of the Indian languages as their medium of communication against 175 million users using English. This gap is predicted to increase by 2021, with users using Indian languages reaching 536 million. Thus, social media companies like Facebook and Internet search companies like Google have increased their support for popularly used Indian languages. Since Hindi is the most widely spoken Indian language, followed by Bengali and Telugu, the introduction of BHAAV dataset is apt and timely.

Related to the task of emotion analysis in Hindi, previous attempts have been made in developing corpus for predicting emotions from Hindi-English code switched language used in social media (Vijay et al., 2018) (*2,866 sentences*), and from auditory speech signals (Koolagudi et al., 2011). Some work has been undertaken in a closely related task of sentiment analysis and datasets have been created for identifying sentiments expressed in movie reviews (Mittal et al., 2013) (*664 reviews*), Hindi blogs (Arora, 2013) (*250 blogs*), and generating lexical resources like Hindi SentiWordnet (Joshi et al., 2010). Given the dearth of resources for analyzing emotions from Hindi text, we present and publicly share **BHAAV**, a corpus of 20,304 sentences collected from 230 different short stories (*e.g.*, *Eidgah* (ईदगाह) by Munshi Premchand) written in Hindi, spanning across 18 genres (see Table 7 for complete list). Each sentence has been annotated by three native Hindi speakers who has at least ten years of reading and writing experience in Hindi language, with the goal of identifying one of the following five popular emotion categories: *anger*, *joy*, *suspense*, *sad*, and *neutral*.

---

[*]Author participated in this research as an adjunct faculty at IIIT-Delhi, India.

[1] https://doi.org/10.5281/zenodo.3457467
[2] https://en.wikipedia.org/wiki/List_of_languages_by_number_of_native_speakers
[3] https://assets.kpmg.com/content/dam/kpmg/in/pdf/2017/04/Indian-languages-Defining-Indias-Internet.pdf

Stories are a melting pot of different types of emotions expressed by the author through the characters and plots that he develops in his writing. Emotions in storytelling has been previously studied resulting in identification of six basic types of emotional arcs in English stories (Reagan et al., 2016), namely - *Rags to riches*, *Tragedy*, *Man in a hole*, *Icarus*, *Cinderella*, and *Oedipus*. This motivated us to develop BHAAV from Hindi stories. We believe that apart from studying emotions in Hindi text, the presented corpus would also enable studies related to the analysis of Hindi literature from the perspective of identifying the inherent emotional arcs. It also has the potential to catalyze research related to human text-to-speech systems geared towards improving automated storytelling experiences. For instance, inducing emotion cues while automatically synthesizing speech for stories from text. We keep the order of the sentences intact as they occur in their source story. This makes the corpus ideal for performing temporal analysis of emotions in the stories, and provides enough information for training machine learning models that takes into account temporal context.

Major contributions of this work are:
- Publicly share the first and the largest annotated Hindi corpus (BHAAV) for sentiment analysis, consisting of 20,304 sentences from 230 popular Hindi short stories spanning across 18 popular genres. Each sentence is labeled with one of the five emotion categories: anger, joy, suspense, sad, and neutral.
- Describe potential applications of BHAAV, the process of annotation, and main challenges in creating an emotion analysis text corpus for a low-resource language like Hindi.
- Propose strong baseline classifiers and report their results for identifying emotion expressed in a sentence of a story written in Hindi.

## 2 Related Work

It is necessary to mention that there has been extensive work in sentiment analysis especially in the past two decades. Although, there is significant intersection between techniques used for sentiment analysis and emotion analysis, yet the two are different in many ways. Emotion analysis is often tackled at a fine grained level and has historically proved to be more challenging due to subtleties involved in identifying and defining emotions. Additionally, resources for emotion analysis are scarce when compared to sentiment analysis, especially for low resource languages like Hindi. For a detailed survey of methods, datasets and theoretical foundations on sentiment analysis and emotion analysis, please refer (Yadollahi et al., 2017; Lei et al., 2018; Cambria et al., 2017; Poria et al., 2017).

Analyzing emotions from text has been primarily manifested through four different types of tasks - *Emotion Detection* (Gupta et al., 2013), *Emotion Polarity Classification* (Alm et al., 2005), *Emotion Classification* (Yang et al., 2007), and *Emotion Cause Detection* (Gao et al., 2015). The scope of this work is limited to the task of *Emotion Classification*. (Pang et al., 2008), mentions that emotions are expressed at four levels - *morphological*, *lexical*, *syntactic* and *figurative*, and noted that as we move from morphological to figurative, the difficulty of the emotion analysis task increases and number of resources for the same decrease. Developed from stories written in a morphologically rich language, BHAAV pri-marily deals with the first and the last levels.

Most of the work for creating data resources for emotion analysis has been fairly limited to building emotion lexicons (Strapparava et al., 2004; Pennebaker et al., 2001; Shahraki and Zaiane, 2017; Mohammad and Turney, 2013), or concentrated in annotating emotions of individual sentences without giving any context (Strapparava and Mihalcea, 2007). As indicated by many, this approach is a non-holistic for a task such as emotion analysis (Schwarz-Friesel, 2015; Ortony et al., 1987). To this end, BHAAV not only presents annotated sentences, but also provides their context.

Lastly, when it comes to the task of analyzing emotions from text, there are no datasets available in Hindi. Although, resource-poor Indian languages have started catching up their richer counterparts in the domain of sentiment analysis (SA) (Mittal et al., 2013; Arora, 2013; Joshi et al., 2010), yet sufficient work needs to be done considering the pace at which these languages are finding their uses in modern digitally driven India. The lack of resources can be judged from the *wide* usage of one of the very few Hindi datasets for SA tasks (Balamurali et al., 2012). It consists of just 200 positive and negative sentences for two major Indian languages, Hindi and Marathi. Another popular and a recent attempt is by (Patra et al., 2015). Their dataset contains approximately 1500 tweets for languages of Hindi, Bengali and Tamil annotated for the task of Aspect Based Sentiment Analysis. BHAAV is certainly an attempt to fill this gap and create a large, effective and high quality resource for emotion mining from text.

## 3 Language Specific Challenges

As already mentioned and pointed in (Yadollahi et al., 2017), the computational methods used in the tasks pertaining to sentiment analysis (SA) can readily be applied to the emotion analysis (EA) tasks. Therefore, the challenges for EA from text are very similar to that of the domain of SA from text. For a detailed description of the challenges one can refer (Mohammad, 2017). However, our task of identifying emotions from sentences poses additional challenges due to the inherent characteristics of Hindi language. We point out some of these language specific challenges (Arora, 2013), in order to draw a complete picture of the intricacies of the task and emphasize that there is a scope of developing methods specific to Hindi, and not all methods developed for English can be directly translated to Hindi.

**Word Order** - The order in which words appear in a sentence plays an important role in determining polarity as well as subjectivity of the text. As opposed to English, which is a *fixed order language*, Hindi is a *free order language*. For any sentence in English to be grammatically correct the 'subject' (S) is followed by 'verb' (V), which is followed by 'object' (O), *i.e.*, in the [SVO] pattern. For example the English sentence - "*Ram (राम) ate (खाया) three mangoes (तीन आम)*", which follows [SVO], can be expressed in the following three ways in Hindi that do not adhere to the [SVO] pattern: (i) 'राम ने तीन आम खाया' [SVO], (ii) 'तीन आम खाया राम ने' [OVS], and (iii) 'खाये तीन आम राम ने' [VOS]. This lack of order can pose challenges to the machine learning algorithms that take into account the order of the words.

**Morphological Variations** - Hindi language is morphologically rich. This means that a lot more information can be expressed in a word in Hindi for which one might end up writing many more words in English. One of the example is that of expressing genders. For example, when using the word 'खायेगी', which means 'will eat' in English, one can not only indicate that someone will eat but also provide cues of the person's gender (in this case female - the male variant is 'खायेगा').

**Handling Spelling Variations** - A word with the same meaning can appear with multiple spelling variations. Occurrence of such variations can pose challenges for the machine learning models that has to take into account all the spelling variants. For example the word 'मेहेंगा', which means 'costly' has another variant महंगा that means the same.

**Lack of Resources** - The lack of lexicons, developed techniques and elaborate resources in Hindi also adds to the challenge, which is also one of the main motivations for our work.

## 4 Corpus Creation and Annotation

One of our primary aims was to create a manually annotated large corpus for performing emotion analysis from text in Hindi. We also wanted to capture the context in which a given piece of text occurs. Therefore, we decided to extract all the sentences from short stories belonging to genres popular in Hindi. Whenever possible we also searched for an audio book[4] where the same story has been narrated by a narrator. This was done in order to help the annotators during the annotation process, in case they have to refer to examples of how a narrator/reader would express the emotion of a sentence in the context of the story. All our annotators were native Hindi speaking volunteers who had a minimum of 10 years of formal education in Hindi, and showed great interest in reading the stories.

| Emotion Category | $\kappa$ | $\alpha$ | Emotions expressed by the category |
|---|---|---|---|
| joy | 0.821 | 0.821 | joy, gratitude, happiness, pleasantness |
| anger | 0.807 | 0.807 | anger, rage, disgust, irritation |
| suspense | 0.757 | 0.757 | wonder, excitement, anxious uncertainty |
| sad | 0.835 | 0.835 | sadness, dis-consolation, loneliness, anxiety, misery |
| neutral | 0.789 | 0.788 | None of the above |
| BHAAV dataset | 0.802 | 0.802 | |

Table 1: Emotions and thier inter-annotator agreements as measured using Fleiss' Kappa ($\kappa$) (Fleiss and Cohen, 1973) and Krippendorff's alpha ($\alpha$) (Krippendorff, 2011) for the entire BHAAV dataset.

The extracted text from 230 stories was split into sentences in an automated way and contained many unnecessary text that were not a part of the story. During the annotation process, the annotators filtered the unwanted text and only annotated the relevant portion. Whenever the sentences were not correctly split, the annotators also corrected them. A total of five annotators were used for annotating the entire corpus, such that each sentence gets at-least three annotations. During the annotation process the annotators had access to the actual online story and the list of audio books. Each story was annotated in one sitting. It took **nine** months to finish the process.

The guidelines for annotating emotions were designed to be very short and concise with regards to the definitions of the categories to be assigned. Due to space restrictions, the guidelines for identifying each emotion are presented in the Appendix. In order to identify emotion categories best suited for our short story corpus, we did some initial annotations with (Plutchik, 1984)'s 'basic' emotions. From the output of the initial phase, we observed that not all basic emotions occurred prominently in the selected Hindi stories. There were five main categories of emotions which were found to be present extensively in the corpus - *anger*, *joy*, *suspense*, *sad*, and *neutral*. A brief description of all the emotion categories is presented in Table 1. A few examples to illustrate the various categories as annotated by the annotators are also given in Table 2. More examples along with common error cases are listed in the Appendix section of the paper.

---

[4]Example of audio books for some of the stories - https://www.youtube.com/user/sameergoswami/playlists

The annotators were instructed not to be biased by their own interpretations of a statement in the story while labeling them. For example, take the case of the following sentence एक दिवसीय क्रिकेट मैच में भारत से हार गया पाक (*Pakistan lost to India in One Day International*). An Indian annotator is often inclined to mark it as *joy* while a Pakistani annotator often marks it as *sad* where as an unbiased reader would read it as having *neutral* emotion. Thus, the annotators were asked to identify only the emotion that an unbiased narrator/reader of that story would like to express while reading it to someone. Whenever confused, they were asked to do the following: first, mark the reason why they think a sentence should have a particular emotion; second, to refer to the audio book of the story if available and try to infer the emotion being expressed; third, if any of the other options do not work, mark it as *neutral*.

General statistics of the dataset are presented in Table 3. As can be seen from the table, Bhaav is imbalanced towards neutral sentences. This is due to the fact that we took raw, unedited stories, making our dataset mimic the distribution of emotions as expressed in the author's writings. Alternatively, in order to balance the dataset, we could have taken selective sentences. However, there would have been several drawbacks associated with such an approach: 1) loss of immediate sentence contexts; 2) separating the individual sentences from the bigger picture as developed by the author in different plots of the story, and 3) failure to capture the implicit emotions expressed by a character of the story (the emotions which a character is feeling *vs* what his words indicate). The overall inter-annotator agreements and the agree-ments for individual emotion categories are presented in Table 1. Next, we present some of the challenges that we faced during the annotation process that we think should be explicitly pointed out in order to provide a true picture of the corpus as well as to give an idea of the difficulties in carrying out such a process.

| Emotion | Sample Hindi Sentences | English Translation |
|---|---|---|
| joy | बादशाह ने कहा तुम्हारी कहानी पहली दोनों से अधिक मनोरंजक है | The king said that your story is more entertaining than the previous two stories |
| anger | रुपिया नई देगा तो उसका खाल उतारकर बाजार में बेच देगा | If he does not give the money then I will take out his skin and sell it in the market |
| suspense | मजदूर ने अब तक तो झलक भर देखी थी अब तो उसे पूरी नजर भर देखा तो ठगा सा खड़ा रह गया | Till now the worker had only seen his glimpses, but when he saw him fully he was just stunned |
| sad | उसने रुँआसे होते हुए मम्मी की ओर देखा | With teary eyes he saw his mother |
| neutral | मैं इसकी मां हूं | I am his mother |

Table 2: Sample sentences from BHAAV dataset for each emotion label.

### 4.1 Challenges in Annotation

Apart from the challenge of annotating a low-resource language for which one can seldom get high quality crowd workers, there were certain challenges that were both specific to the domain of stories as well as generic ones peculiar to the tasks of sentiment and emotion analysis. Some of the prominent ones as identified from the feedback of the annotators are presented below with examples.

**Identifying Implicit Emotions** - The annotators were asked to identify the emotions whenever it was both explicitly and implicitly expressed. An example of explicitly expressed emotion would be - Example 1, in which the speaker by using the words such as सुहावना (refreshing), मनोहर (beautiful) clearly indicates that he is happy with the nature, thus expressing his joy in the statements.

. *Example 1* - कितना मनोहर, कितना सुहावना प्रभाव है| वृक्षों पर अजीब हरियाली है, खेतों में कुछ अजीब रौनक है, आसमान पर कुछ अजीब लालिमा है| *(It is such a beautiful and enjoyable feeling. There is a strange greenery on the trees, some strange liveliness in the fields, there is some weird but enjoyable redness in the sky)*

Identifying implicit emotions were sometimes confusing for the annotators and on taking a closer look we did find some of them being marked as neutral. An example of implicitly expressed emotion would be - Example 2, in which a child's grandmother is complaining about her son being too hasty of going to the mosque. She complains of his ignorance of knowing anything about driving a household and its inherent difficulties. Although there are no explicit words indicating her state of the mind, there is an implicit pointer that she is feeling irritated due to the haste and hence is angry over him. These types of emotions are totally contextual and could be identified only while reading the story. We believe that capturing these emotions are also necessary in order to make our annotation process holistic. Although, we do not train any classification model in this work that can take these types of context in order to predict the final emotion of a sentence, yet we think that BHAAV as a dataset provides an opportunity to build such contextual models making it a rich corpus unlike many other previous ones as already pointed out in Section 2. We would certainly like to take it up as a future work.

. *Example 2* - अब जल्दी पड़ी है कि लोग ईदगाह क्यों नहीं चलते| इन्हें गृहस्थी की चिंताओं से क्या प्रयोजन| *(Now he is feeling why do not people go to the mosque a little faster. What do they (the children) know about household chores)*

**Primary Target of Opinion** - Another challenge comes when there is not even an implicit clue in the immediate context of a sentence. For instance, in a story, sometimes a character is developed as an adversary to a particular prop (*i.e.*, PTO, Primary Target of Opinion). The prop can be another character or some inanimate object or phenomena. From the start of the story, the character expresses his emotions in a characteristic manner towards that PTO. Thus if a sentence or a context does not have any explicit clues to know the state of the mind of the character, identifying the PTO and the character's emotions towards PTO gives some connotation to that sentence. This is in line to what was suggested in the work (Mohammad, 2016). An example of such an instance as presented in Example 3, can be derived from the famous story by Munshi Premchand, *Eidgah*. The following sentence when read in isolation could potentially trick someone into thinking whether the boy speaking these dialogues is expressing mercy or even neutrality, when he is actually expressing joy.

- *Example 3* - मोहसिन- लेकिन दिल में कह रहे होंगे कि मिले तो खा लें| *(Mohsin- But in the hearts, they must be thinking that if they could get it, they would eat it)*

**Sarcasm** - A common challenge which annotators faced while annotating BHAAV is the case of sarcasm, which is again prevalent in most of the previous works in sentiment and emotion analysis. Sarcasm, as it occurs, is generally accompanied by either anger or delight (or sometimes both) of the speaker at the dismay of the PTO. Thus, in most cases, the emotional state of the speaker of sarcastic comments was a mixture of anger with the PTO and rejoicement at its expense. However, to account for the headline categories we chose for Bhaav, annotators were asked to differentiate between these two causes using the context provided and mark the category which most closely represents the sentence. This was sometimes challenging. For instance, in example 4 the emotion most close to the state of speaker is that of *anger*, when it could be easily misunderstood to be that of *joy*.

- *Example 4* - हा हा हा! अब तुम बताओगे हम क्या बोलें? *(Ha Ha Ha ! Now you would tell me what I should speak?)*

**Annotating Suspense** - Suspense was the toughest category for the annotators and proved very difficult for them to know exactly when a sentence is of this category. The annotators were asked to mark a sentence as suspense when there is some element in it which evokes a sense of wonder, anticipation or worry (see Example 5). Suspense is a unique feature of stories which does not get fully expressed in other types of written materials such as news articles, formal reports, and others.

- *Example 5* - पिछले पहर को महफिल में सन्नाटा हो गया| हू-हा की आवाजें बन्द हो गयीं| लीला ने सोचा, क्या लोग कहीं चले गए, या सो गये? एकाएक सन्नाटा क्यों छा गया? *(Last afternoon, the silence was over the entire place. There were no voices around. The sounds of Hu-Ha completely stopped. Leela thought, did people go somewhere, or perhaps they slept? Why all of a sudden there is silence everywhere?)*

Next, we present the experiments performed for training the baseline models.

## 5 Baseline Models

In this section, we describe strong baseline models that we train for the task of one of the emotions - *anger*, *joy*, *suspense*, *sad*, and *neutral*, from a given sentence taken from a Hindi story. Both classic machine learning and modern deep learning models are trained and their results are analyzed. We extensively use Sklearn (Pedregosa et al., 2011) and Keras (Chollet et al., 2018) as our machine learning toolkits.

### 5.1 Dataset

| Emotion | No. of Sentences | No. of Sentences (Train data) | No. of Sentences (Test data) |
|---|---|---|---|
| joy | 2,463 | 2,242 | 221 |
| anger | 1,464 | 1,321 | 143 |
| suspense | 1,512 | 1,389 | 123 |
| sad | 3,168 | 2,843 | 325 |
| neutral | 11,697 | 10,478 | 1,219 |

Table 3: Distribution of sentences in different categories of emotions in the BHAAV dataset.

The BHAAV dataset was randomly shuffled and split into train and test datasets with a ratio of 10:1. The distribution of labels in the two datasets are shown in Table 3. The proportion of distribution of labels in the test dataset is kept similar to the training dataset. We train our models on the training dataset and test the final predictions on the test dataset. We do not create a separate validation dataset. However, we do use validation data extracted from the training data, whenever necessary for tuning the hyperparameters of the models.

### 5.2 Text Preprocessing

Before training the classification models one needs to preprocess the text and represent each sentence as a feature vector. We tokenize each sentence into words and remove punctuations. We do not remove the stopwords. Since we deal with Hindi, the standard word tokenizers that are suitable for English language could not be used. Therefore, we used the tokenizer shipped with Classical Language Toolkit[5]. Each sentence is vectorized after a feature extraction step for the classic machine learning models such as Support Vector Machines. Unigrams, Bigrams and Trigrams were generated as features for each sentence and their TF-IDF (Aizawa, 2003) scores were considered as the feature values.

One of the key components of the input fed to the deep learning models are pre-trained word embeddings (Kusner et al., 2015), that are used for representing each word of the input sentences by a dense real valued vector. Since the dataset on which we train our models is relatively small, we use the pretrained word embeddings in order to prevent overfitting. This practice is commonly known as transfer learning[6]. We choose the Fasttext[7] word embeddings (Bojanowski et al., 2016), trained on the Hindi Wikipedia corpus. This was a natural choice due to its easy availability. Additionally, Fasttext is possibly a better choice than other popular word embedding methods as it is more suitable for representing words belonging to morphologically rich languages like Hindi as described in Section 3.

---
[5]http://docs.cltk.org/en/latest/hindi.html

While training the deep learning models, each sentence in the training and test dataset is converted to a fixed size document of 126 words (maximum length of a sentence in the dataset). Padding[8] is used for sentences of length lesser than 126 words. Each word is represented as a 300 dimensional (D) vector by the word embedding model. All the words in the dataset are mapped to their corresponding word embedding vector. Whenever a word is not found in the vocabulary of the word embedding model we assign it a 300-D zero vector. Each sentence is then represented as a matrix of its constituent words and their corresponding embedding vector, which is then fed as an input to deep learning algorithms.

### 5.3 Training

All the machine learning models were trained after selecting the hyperparameters on a validation data. 10-fold cross validation was used for the classic techniques. For the deep learning models, random search (Bergstra and Bengio, 2012) was used for selecting the best hyperparameters among the ones shown in Table 4, that best fitted a fixed randomly selected validation data comprising of 20% of the training data. Only 100 iterations of random search was performed. Once the hyperparameter tuning was done the final model was trained on the entire training data using the selected hyperparameters. Adam (Kingma and Ba, 2014) with two annealing restarts has been shown to work faster and perform better than SGD in other NLP tasks (Denkowski and Neubig, 2017). Therefore, we use the same as our optimization algorithm for the deep learning models. As the task is a multi-class classification problem, categorical cross entropy was used as the loss function, and the final layer of both the deep learning models consisted of a fully-connected dense neural network with the extracted features as the input and a softmax output giving the prediction probability for each of the five emotion categories.

---

[6] ftp://ftp.cs.wisc.edu/machine-learning/shavlik-group/torrey.handbook09.pdf
[7] https://github.com/facebookresearch/fastText/blob/master/pretrained-vectors.md
[8] https://keras.io/preprocessing/sequence/

| Hyperparameter | Range |
|---|---|
| No. of Filters for CNN | 100, 200, 300, 400 |
| Filter sizes for the CNN model | 1, 2, 3, 4, 5, 6 |
| Dense Output Layer Size | 100, 200, 300, 400 |
| Dropout Probability | 0.2, 0.3, 0.4, 0.5, 0.6, 0.7, 0.8, 0.9 |
| Learning Rate | 0.0001, 0.001 |
| Batch Sizes | 8, 16, 32, 64, 128 |
| Epochs | 10, 50, 100, 150 |
| LSTM units | 8, 16, 32, 64, 128, 256 |

Table 4: Hyperparameter ranges used for random search during training deep learning models (CNN and Bidirectional LSTM).

Among the classic machine learning techniques, *Support Vector Machine* (SVM) with a linear kernel (Hsu et al., 2003), *Logistic Regression* (Yu et al., 2011) and *Random Forests* (Breiman, 2001) were trained. A shallow Convolutional Neural Network with a single input channel similar to (Severyn and Moschitti, 2015), and Bidirectional Long Short Term Memory networks with an architecture similar to (Mahata et al., 2018), are the deep learning models that were trained. A random classifier that randomly generated predictions from a label distribution similar to that of the training dataset was also implemented. Table 5 summarizes the performances of the classifiers on the test dataset for the following metrics - *macro average precision*, *macro average recall*, *macro average F1-score*, and *accuracy* (Sokolova and Lapalme, 2009). We chose macro-average measures as the data is imbalanced and macro-averaging will assign equal weights to all the categories, which gives a better generic performance of any classifier.

## 6 Discussion

In order to analyze the possible features chosen by a machine learning classification algorithm for discriminating between different categories of emotions and to validate the ability of the BHAAV dataset in providing such features to any classifier, we looked at the most important features chosen by the Logistic Regression model. Table 6 shows the top 10 most informative unigram features for each category of emotion chosen by the model in order to make the final predictions. As evident from the choices, words like प्रसन्न (glad), सुंदर (beautiful), खुश (happy), हँस, (laugh), are sensible indicators of *joy*, and so are the words like अपमान (insult), गुस्सा (anger), क्रोध (anger), बदला (revenge), for *anger*. The other categories also show a similar pattern.

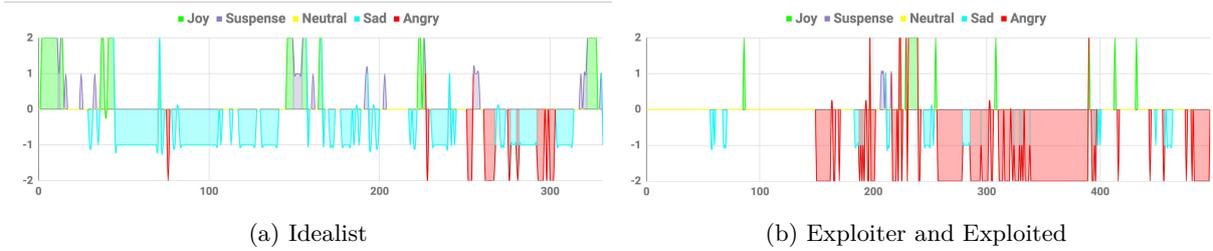

| | (a) Idealist | (b) Exploiter and Exploited |

Figure 1: Flow of emotions in randomly selected stories from two different genres.

| Method | Macro Avg Precision | Macro Avg Recall | Macro Avg F1 | Accuracy |
|---|---|---|---|---|
| **Logistic Regression** | 0.58 | 0.62 | 0.58 | 0.62 |
| **SVM** | 0.48 | 0.52 | 0.49 | 0.52 |
| **Random Forests** | 0.44 | 0.59 | 0.45 | 0.59 |
| **CNN** | 0.50 | 0.55 | 0.51 | 0.55 |
| **BLSTM** | 0.43 | 0.60 | 0.47 | 0.60 |
| **Random Classifier** | 0.40 | 0.40 | 0.40 | 0.40 |

Table 5: Performance of the baseline supervised classification models on BHAAV dataset.

| Emotion | Top 10 Important Unigram Features |
|---|---|
| **joy** | प्रसन्न (glad), सुंदर (beautiful), खुश (happy), हँस, (laugh), संगीत (music), खिलौने (toys), मजा (fun), आनंद (joy), हँसकर (smilingly), उछल (jump) |
| **anger** | अपमान (insult), गुस्सा (anger), क्रोध (anger), बदला (revenge), मूर्ख (idiot), सजा (punishment), जहन्नुम (hell), आग (fire), दुष्ट (evil), चिल्लाया (screamed) |
| **suspense** | आवाज़ (sound), आश्चर्य (astonishment), जिन्न (Genie), देखा (saw), युद्ध (war), छन… (sound of anklets), कहाँ (where), जादू (magic), अचानक (suddenly), जहाज (ship), |
| **sad** | रो (cry), मर (die), रोने (crying), दुख (sadness), हृदय (heart), दुखी (sad), जीवन (life), आँसू (tears), रोते (cry), भगवान् (God) |
| **neutral** | किसान (farmer), उसने (he), बिन्नी (Binny), पूछा (asked), दादाजी (grandfather), कल (tomorrow), पंडित (pundit), मेहता (mehta), मां (mother), आना (come) |

Table 6: Top 10 most important features for each emotion category as identified by the Logistic Regression model during training.

We also looked at the performance of the classifiers for individual categories. *Neutral* category had the best performance consistently, which is quiet easy to guess from the data distribution (Table 3) and it being the majority class. The performance of the *suspense* category was consistently low. Although, the category of *anger* had a similar presence in BHAAV, yet it had better performance than *suspense*. This might be due to the presence of better discriminative features for *anger* than *suspense*. Another reason could be related to challenges associated with annotating the *suspense* category (Section 4.1).

Our analysis provides a brief insight into the BHAAV dataset from which we can conclude that it is an appropriate dataset for emotion identification and classification tasks. Although, the dataset is created from stories, it can possibly be used for many other domains as it is rich in features indicating the five different emotions as presented in this work. The annotations were done from the perspective of a reader/narrator trying to express the emotion of a sentence, given the existing scenario in the story and whenever applicable trying to express the emotion of a character in the story. This also makes this dataset suitable for training automated text-to-speech interfaces (*e.g.*, audio books) for story narration and improving them by infusing emotions in them.

### 6.1 Emotions and Genres

We started with frequently used 30 genres as mentioned by (Nagendra, 1994) and selected 500 popular online short stories. However, we narrowed down to the most frequent 18 genres (see Table 1 for complete list) and ended up with extracting text from 230 stories, depending on the availability of online content. Throughout the process of deciding on genres and finding online content relevant to them, we took help from some experts in Hindi literature who have done their PhD in Hindi liter-ature.

| Genres |
|---|
| आदर्शवादी (Idealist) |
| प्रेमपरक (Romantic) |
| शहरी जीवन (Urban Life) |
| शोषक और शोषित वर्ग (Exploiter and Exploited Class) |
| नीतिपरक (Moral Stories) |
| किसान जीवन (Life of a Farmer) |
| ऐतिहासिक (Historical) |
| प्रेरणादायक (Inspiration) |
| देश भक्ति संबंधित (Patriotic) |
| व्यक्तिगत जीवन की समस्या (Personal Issues/Problems) |
| रूढ़ि और अंधविश्वास (Dogmatic and Superstitious) |
| संयुक्त परिवार की समस्या (Joint Family Problems) |
| रहस्यमयी (Mystery) |
| यथार्थवादी (Realistic and Pragmatic) |
| ग्रामीण (Village Life) |
| उपदेशपरक (Instructive) |
| भोगे हुए यथार्थ की कहानी (Real Stories) |
| समाज सुधारक (Society and its Reformation) |

Table 7: Genres present in BHAAV

BHAAV is appropriate for analyzing the flow of emotions in individual stories and study them for different genres. We plotted the flow of emotions in a randomly picked story from two different genres as shown in Figure 1. It is observable from the figures that each story has its own distinct emotion footprint. It would be interesting to study them and draw interesting linguistic insights from the Hindi literature using BHAAV.

## 7 Future Work and Conclusion

In this work we publicly shared the first and the largest annotated corpus, named BHAAV, with 20,304 sentences in Hindi, for emotion analysis. We provided a detailed description of the dataset, language specific challenges, annotation process, challenges associated with annotations and reported performances of the baseline classification models trained on the dataset for identifying emotions expressed in a sentence. Through different observations we confirm BHAAV to be rich with emotion cues and point to the potential applications. In the future, we plan to work on enriching BHAAV with more annotations related to sentiment and discourse analysis, and believe that it will prove to be a valuable resource in Hindi.

## A  General Instruction

- Attempt HITs only if you are a native speaker of Hindi.

- Your responses are confidential. Any publications based on these responses will not include your specific responses, but rather aggregate information from many individuals.We will not ask any information that can be used to identify who you are.

## B  Task Specific Instructions

- We take into account these five headline categories: Anger, Joy, Sad, Suspense, Neutral/ Plain Talk.

- The headline and subordinate categories are as mentioned below

    - **Anger(0)** - Emotions include anger, rage, disgust, violent unwillingness, sadism, irritation
    - **Joy(1)** - Emotions include Joy, gratitude, happiness, pleasantness, elation, positive excitement, triumph, gratification, pride
    - **Sad(2)** - Emotions include sadness, disconsolation, loneliness, anxiety, misery, sorry, depressing, shameful, grief-stricken, melancholy, unwilling
    - **Suspense(3)** - Wonder, excitement, anxious uncertainty
    - **Neutral(4)** / Plain talk - These include no emotions, examples are general talk spoken with no emotion

- Agreeing or disagreeing with the speaker's views should not have a bearing on your response. You are to assess the language being used (not the views). For example, given the tweet, '*Evolution makes no sense*', the correct answer is '*the speaker is using negative language*' since the speaker's words are criticizing or judging negatively something (in this case the theory of evolution). Note that the answer is not contingent on whether you believe in evolution or not.

- From reading the text, identify the entity towards which opinion is being expressed or the entity towards which the speaker's attitude can be determined. This entity is usually a person, object, company, group of people, or some such entity. We will call this the PRIMARY TARGET OF OPINION (PTO). For example, if the text criticizes certain actions or beliefs of a person (or group of persons), then that person or group is the PTO. If the text mocks people who do not believe in evolution, then the PTO is 'people who do not believe in evolution'. If the text questions or mocks evolution, then the PTO is 'evolution'.

- While annotating, always try to find an explicit or implicit clue which suggests the speakers' attitude towards the situation. The speaker in this reference can be the narrator himself or the characters of the story. Example of a clue can be positive words or sentiments described in a sentence explaining a situation.

  *Example* - रमजान के पूरे तीस रोजों के बाद ईद आयी है | कितना मनोहर, कितना सुहावना प्रभाव है | वृक्षों पर अजीब हरियाली है, खेतों में कुछ अजीब रौनक है, आसमान पर कुछ अजीब लालिमा है |

  Here, the narrator narrates the story about a month Ramzan. The narrator though could be plain talking but the sentence, "कितना मनोहर, कितना सुहावना प्रभाव है" give us a clue that he is not simply stating the events as is. Rather, he has an emotional attachment to the climate and the story settings. In particular, he is happy about the environment and its refreshing events. Thus, using this clue we can know for sure, that these sentences are not neutral but contain an emotion of joy.

- In case where someone is just quoting another person with no reference to his own emotional state, find explicit or implicit clue which suggests the speaker's attitude towards PTO.

## C  Example HIT

- **Joy**: रमजान के पूरे तीस रोजों के बाद ईद आयी है | कितना मनोहर, कितना सुहावना प्रभाव है | वृक्षों पर अजीब हरियाली है, खेतों में कुछ अजीब रौनक है, आसमान पर कुछ अजीब लालिमा है | आज का सूर्य देखो, कितना प्यारा, कितना शीतल है, यानी संसार को ईद की बधाई दे रहा है | गाँव में कितनी हलचल है | ईदगाह जाने की तैयारियाँ हो रही हैं | *(Eid has come after 30 days of Ramadan. It is such a beautiful and enjoyable feeling. There is a strange greenery on the trees, some strange liveliness in the fields, there is some weird but enjoyable redness in the sky. Look at today's sun is looking, how lovely, how cool it is, that is to congratulate the world on Eid. There is so much commotion in the village? Preparations are going to go to Idgah.)*

  Here the narrator is expressing his joy towards the change in season and the coming of the festival.

- **Anger**: लड़के सबसे ज्यादा प्रसन्न हैं | किसी ने एक रोजा रखा है, वह भी दोपहर तक, किसी ने वह भी नहीं, लेकिन ईदगाह जाने की खुशी उनके हिस्से की चीज है | रोजे बड़े-बूढ़ों के लिए होंगे | इनके लिए तो ईद है | रोज ईद का नाम रटते थे, आज वह आ गयी | अब जल्दी पड़ी है कि लोग ईदगाह क्यों नहीं चलते | इन्हें गृहस्थी की चिंताओं से क्या प्रयोजन | सेवैयों के लिए दूध और शक्कर घर में है या नहीं, इनकी बला से, ये तो सेवेयाँ खायेंगे | वह क्या जानें कि अब्बाजान क्यों बदहवास चौधरी कायमअली के घर दौड़े जा रहे हैं | उन्हें क्या खबर कि चौधरी आँखें बदल लें, तो यह सारी ईद मुहर्रम हो जाय | *The boys are most pleased. Someone has kept a rosa, that too by noon, someone hasn't, but the joy of going to Idgah is the part of their share. Rose will be for the elderly. For them it is Eid. Every day the people use to talk about Eid, today it came. Now they are excited, asking why do not people go to the mosque a little faster. What do they (the children) know about household chores? .They are not bothered whether there is milk or sugar in the house for saivanya (a type of food), they just want to eat it. What does he know why the father is going to Chowdhary (ask for money to celebrate Eid). They don't know if Chowdhary changes his mood, Eid would become Muharram.*

  In the last three sentences, the narrator shows signs of irritation, which is a subcategory of the headline category, "Anger".

- **Suspense**: पिछले पहर को महफिल में सन्नाटा हो गया | हू-हा की आवाजें बन्द हो गयीं | लीला ने सोचा,

क्या लोग कहीं चले गए, या सो गये | एकाएक सन्नाटा क्यों छा गया | *(Last afternoon, the silence was over the entire place. There were no voices around. The sounds of Hu-Ha completely stopped. Leela thought, did people go somewhere, or perhaps they slept? Why all of a sudden there is silence everywhere?)*

Here the narrator is trying to create suspense.

- **Sad**: उन्होंने खुद वह सब कष्ट झेले हैं, जो वह मुझे झेलवाना चाहती हैं | उनके स्वास्थ्य पर उन कष्टों का जरा भी असर नहीं पड़ा | वह इस 65 वर्ष की उम्र में मुझसे कहीं टाँठी हैं | फिर उन्हें कैसे मालूम हो कि इन कष्टों से स्वास्थ्य बिगड़ सकता है | *(She herself has experienced all the hardships and she wants me to do the same. Those sufferings did not have any effect on her health. She is much healthier than me despite being 65 years of age. Then how does she will come to know that health problems are worsened by these sufferings?)*

Once the wife justifies her mother-in-law's actions, she starts explaining herself and her deplorable situation by taking the reference of her failing health. Due to the continuous arguments and hardships she has to face due to her mother-in-law, her health is suffering. She even says her mother-in-law at the age of 65 is healthier than herself. Thus, here she is showing the signs of her being sad about her situation. Thus, this comes under the category 'Sad'.

- **Neutral**: गाँव से मेला चला | और बच्चों के साथ हामिद भी जा रहा था | कभी सबके सब दौड़कर आगे निकल जाते | फिर किसी पेड़ के नीचे खड़े होकर साथ वालों का इंतज़ार करते | *(A group of people from the village left for Idgah. And with the kids, hammid was also going. Sometimes they start running in an attempt to outdo the others. Then they stood under a tree and waited for the people to catch up to them.)*

Here although the sentence itself is positive but the narrator is not emotionally attached to the situation. He is just reporting it as is. There is no clue whatsoever which indicates that the narrator is expressing a positive sentiment with the reportage.